\pdfoutput=1

\documentclass[unnumsec,webpdf,modern,large]{oup-authoring-template}

\onecolumn 

\graphicspath{{Fig/}}

\theoremstyle{thmstyleone}%
\theoremstyle{thmstyletwo}%
\theoremstyle{thmstylethree}%

\usepackage{natbib} 
\setcitestyle{square,comma,numbers}
\usepackage{graphicx}
\usepackage{booktabs}

\newcommand{\eat}[1]{\ignorespaces}

\newcounter{hoifung}
\newcommand{\hoifung}[1]{\refstepcounter{hoifung}{\todo[color=red!40, size=\footnotesize]{[\textbf{hoifung:\thehoifung}] #1}}}

\newcounter{rob}
\newcommand*{\rob}[1]{%
\refstepcounter{rob}%
{%
\todo[color=blue!40, size=\footnotesize]{%
[\textbf{rob:\therob}] #1}%
}}%

\newcommand{\sam}[1]{\ignorespaces}

\newcommand{\munotes}[1]{\ignorespaces}

\newcommand{\paul}[1]{\ignorespaces}

\newcommand{\brian}[1]{\ignorespaces}

\begin{document}

\journaltitle{Preprint}
\DOI{DOI HERE}
\copyrightyear{2022}
\pubyear{2022}
\access{Advance Access Publication Date: Day Month Year}
\appnotes{Research and Applications}

\firstpage{1}


\title[Towards Structuring Real-World Data at Scale]{Towards Structuring Real-World Data at Scale: Deep Learning for Extracting Key Oncology Information from Clinical Text with Patient-Level Supervision}

\author[1,$\ast$]{Sam Preston}
\author[1,$\ast$]{Mu Wei}
\author[1]{Rajesh Rao}
\author[1]{Robert Tinn}
\author[1]{Naoto Usuyama}
\author[1]{Michael Lucas}
\author[2]{Roshanthi Weerasinghe}
\author[2]{Soohee Lee}
\author[3]{Brian Piening}
\author[3]{Paul Tittel}
\author[1]{Naveen Valluri}
\author[1]{Tristan Naumann}
\author[3,$\ast\ast$]{Carlo Bifulco}
\author[1,$\ast\ast$]{Hoifung Poon}

\authormark{Preston, Wei et al.}

\address[1]{
    \orgname{Microsoft Research}, 
    \orgaddress{
        \street{One Microsoft Way},
        \state{Redmond, WA},
        \postcode{98052},  
        \country{USA}
    }
}
\address[2]{
    \orgdiv{Providence Research Network},
    \orgname{Providence St Joseph’s Health},
    \orgaddress{
        \state{Portland, OR},
        \postcode{97213},
        \country{USA}
    }
}
\address[3]{
    \orgname{Providence Genomics \& Earle A. Chiles Research Institute},
    \orgaddress{
        \street{4805 NE Glisan}, 
        \state{Portland, OR},
        \postcode{97213},
        \country{USA}
    }
}

\corresp[$\ast$]{These authors contributed equally to this work.\\}
\corresp[$\ast\ast$]{Corresponding authors: \href{email:Carlo.Bifulco@providence.org}{Carlo.Bifulco@providence.org} and \href{email:hoifung@microsoft.com}{hoifung@microsoft.com}}

\received{Date}{0}{Year}
\revised{Date}{0}{Year}
\accepted{Date}{0}{Year}



%

\abstract{
\textbf{Objective:} The majority of detailed patient information in real-world data (RWD) is only consistently available in free-text clinical documents. Manual curation is expensive and time-consuming. Developing natural language processing (NLP) methods for structuring RWD is thus essential for scaling real-world evidence generation.\\
\textbf{Materials and Methods:} Traditional rule-based systems are vulnerable to the prevalent linguistic variations and ambiguities in clinical text, and prior applications of machine-learning methods typically require sentence-level or report-level labeled examples that are hard to produce at scale. 
We propose leveraging patient-level supervision from medical registries, which are often readily available and capture key patient information, for general RWD applications. 
To combat the lack of sentence-level or report-level annotations, we explore advanced deep-learning methods by combining domain-specific pretraining, recurrent neural networks, and hierarchical attention.\\ 
\textbf{Results:} We conduct an extensive study on 135,107 patients from the cancer registry of a large integrated delivery network (IDN) comprising healthcare systems in five western US states. Our deep learning methods attain test AUROC of 94-99\% for key tumor attributes and comparable performance on held-out data from separate health systems and states.\\ 
\textbf{Discussion and Conclusion:} Ablation results demonstrate clear superiority of these advanced deep-learning methods over prior approaches. Error analysis shows that our NLP system sometimes even corrects errors in registrar labels. 
We also conduct a preliminary investigation in accelerating registry curation and general RWD structuring via assisted curation for over 1.2 million cancer patients in this healthcare network.
}

\keywords{Natural Language Processing (L01.224.050.375.580); Data Mining (L01.313.500.750.280.199); Medical Oncology (H02.403.429.515); Neoplasm Staging (E01.789.625)}

\eat{

We propose leveraging patient-level supervision from medical registries, 
which are often readily available and capture key patient information, 
for general RWD applications. 


}


\maketitle

\eat{
JAMIA: Word count: up to 4000 words.
Structured abstract: up to 250 words.
Tables: up to 4.
Figures: up to 6.
}

%

\section{INTRODUCTION}

Electronic medical records (EMRs) offer an unprecedented opportunity to harness real-world data (RWD) for accelerating progress in clinical research and care~\cite{butteRWD2020}. 
By tracking longitudinal patient care patterns and trajectories, including diagnoses, treatments and clinical outcomes, we can help assess drug efficacy in real-world settings, facilitate post-market surveillance, and speed up clinical trial recruitment. However, pertinent information about patients often resides in clinical text, such as pathology assessments, radiology assessments, and clinical progress notes. Manual curation to structure such text is expensive and hard to scale.


Natural language processing (NLP) can help accelerate manual curation~\cite{luAssistedCuration2012}. 
In recent years, there have been rapid advances in general-domain NLP, where state-of-the-art deep neural networks, such as transformer-based models, have demonstrated remarkable success across a wide range of applications~\cite{devlin2019bert,liu2019roberta}. 
Training these sophisticated models, however, typically requires a large number of annotated examples. By contrast, prior work in clinical NLP is often limited to annotating small datasets 
and training simpler methods~\cite{kehl2019assessment}.
Due to the scarcity of qualified domain experts, annotation is usually conducted on a small collection of notes, often from a single institution. Moreover, to make learning easier, these explorations typically restrict annotation to single sentences or single notes.
For example, {Kehl et al.~\cite{kehl2019assessment}} shows promising results on applying NLP to accelerate real-world evidence generation in oncology. However, while their annotation effort is relatively large among similar prior efforts, their test set contains only 109 patients (1,112~patients in the entire annotated dataset). The notes are limited to radiology reports for lung cancer from a single institution. Their exploration is limited to convolutional neural networks, which does not leverage the latest NLP advances, such as language model pretaining~\cite{oliver&al21,gu2020domain}.

In this paper, we propose to bootstrap deep learning for structuring real-world data (RWD) by using readily available registry data. Medical registries are routinely collected for various diseases, with oncology being a prominent example. In the U.S., cancer is a reportable disease and cancer centers are required to curate patient information per national accreditation and clinical quality requirements. By matching registry entries with their corresponding EMR data, we can assemble a large dataset for training and evaluating state-of-the-art deep NLP methods. 

{Gao et al.~\cite{gao2021limitations,gao2019classifying}} also leverage registry data for supervision. However, like {Kehl et al.~\cite{kehl2019assessment}}, they restrict classification to individual pathology reports, and exclude tumors associated with multiple reports. 
Similarly, {Percha et al.~\cite{percha21curation}} focuses on classifying individual sentences for breast cancer surgery information. 
Such methods are not applicable to the prevalent cases where information is scattered across multiple clinical documents and note types (e.g., pathology reports, radiology reports, progress notes). Often, information in a single document (e.g. discussion of a malignant site) is insufficient and additional context is required for identifying the correct diagnosis or staging information. 

To the best of our knowledge, our study is the first to explore cross-document medical information extraction, using registry-derived patient-level supervision to train deep NLP methods. 
Such patient-level supervision is inherently challenging to use as it comprises only annotations associated with a tumor diagnosis, which are not attributable to individual sentences or documents. 
Each patient may have dozens of clinical documents, yielding very long input text spans that are difficult to process for standard deep learning methods. 
Additionally, the collection of clinical documents spans decades and varies in completeness. Nevertheless, we found that the scale of such self-supervised data more than compensates for its noise and technical challenges, and our models attain high performance (AUROC 94-99\%) for extracting core tumor attributes such as site, histology, clinical/pathological staging. 

Unlike in simplistic settings of prior studies~\cite{gao2021limitations}, sophisticated deep learning methods substantially outperform simplistic approaches, with our top-performing model combining cutting-edge techniques such as transformers~\cite{devlin2019bert}, domain-specific pretraining~\cite{gu2020domain}, recurrent neural networks~\cite{cho2014learning}, and hierarchical attention~\cite{yang2016hierarchical}. 
Our method naturally handles longitudinal information and experiments show that incorporating multiple document types significantly improves performance. Neural attention can be used to pinpoint relevant text spans as extraction rationale and provenance, which facilitate model interpretation and rapid validation by human experts. Our model trained on a health system in one state performed comparably for patients from different states, health systems and EMR configuration, suggesting good generalizability. 


While our work is motivated by structuring real-world data, our method can also be used to accelerate registry curation. Our deep learning model not only performs well in abstraction, but also attains high accuracy in case finding (identifying patients for cancer registry), thus paving the way for end-to-end assisted cancer registry curation.

\eat{
\hoifung{TODO: add refs below}
<< TODO ADD REFS BELOW >>

\cite{hao-etal-2020-enhancing}: Enhancing Clinical BERT Embedding using a Biomedical Knowledge
Base / Boran Hao and Henghui Zhu and Ioannis Paschalidis, COLING 2020

\cite{zhu21distantly}: Distantly supervised biomedical relation extraction using piecewise attentive convolutional neural network and reinforcement learning. Tiantian Zhu, Yang Qin, Yang Xiang, Baotian Hu, Qingcai Chen, Weihua Peng / JAMIA 2021, Volume 28, Issue 12

\cite{meystre&al22}: Natural language processing enabling COVID-19 predictive analytics to support data-driven patient advising and pooled testing
Stéphane M Meystre, Paul M Heider, Youngjun Kim, Matthew Davis, Jihad Obeid, James Madory, Alexander V Alekseyenko
Journal of the American Medical Informatics Association, Volume 29, Issue 1, January 2022, Pages 12–21

\cite{oliver&al21}: Clinically relevant pretraining is all you need
Oliver J Bear Don’t Walk IV, Tony Sun, Adler Perotte, Noémie Elhadad
Journal of the American Medical Informatics Association, Volume 28, Issue 9, September 2021, Pages 1970–1976

\cite{percha21curation}: Natural language inference for curation of structured clinical registries from unstructured text
Bethany Percha, Kereeti Pisapati, Cynthia Gao, Hank Schmidt
Journal of the American Medical Informatics Association, Volume 29, Issue 1, January 2022, Pages 97–108
}
\section{MATERIALS AND METHODS}

\begin{figure}[t]
    \centering
    \includegraphics[width=0.9\textwidth]{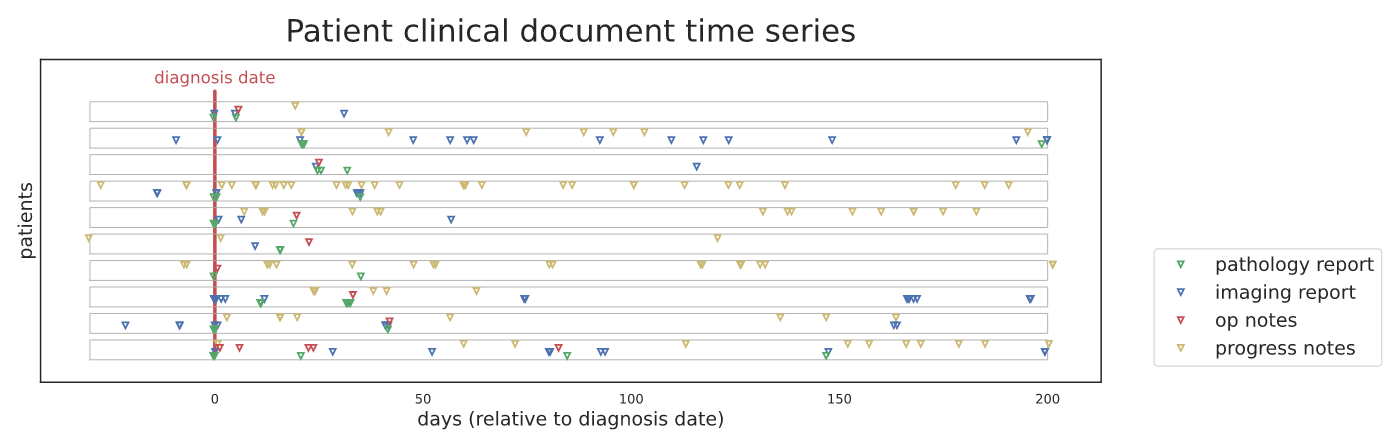}
    \caption{
    Cancer patients typically have many clinical documents for a tumor diagnosis, with key information scattered around.
    }
    \label{fig:multiple-notes}
\end{figure}

\begin{figure}[t]
    \centering
    \includegraphics[width=0.9\textwidth]{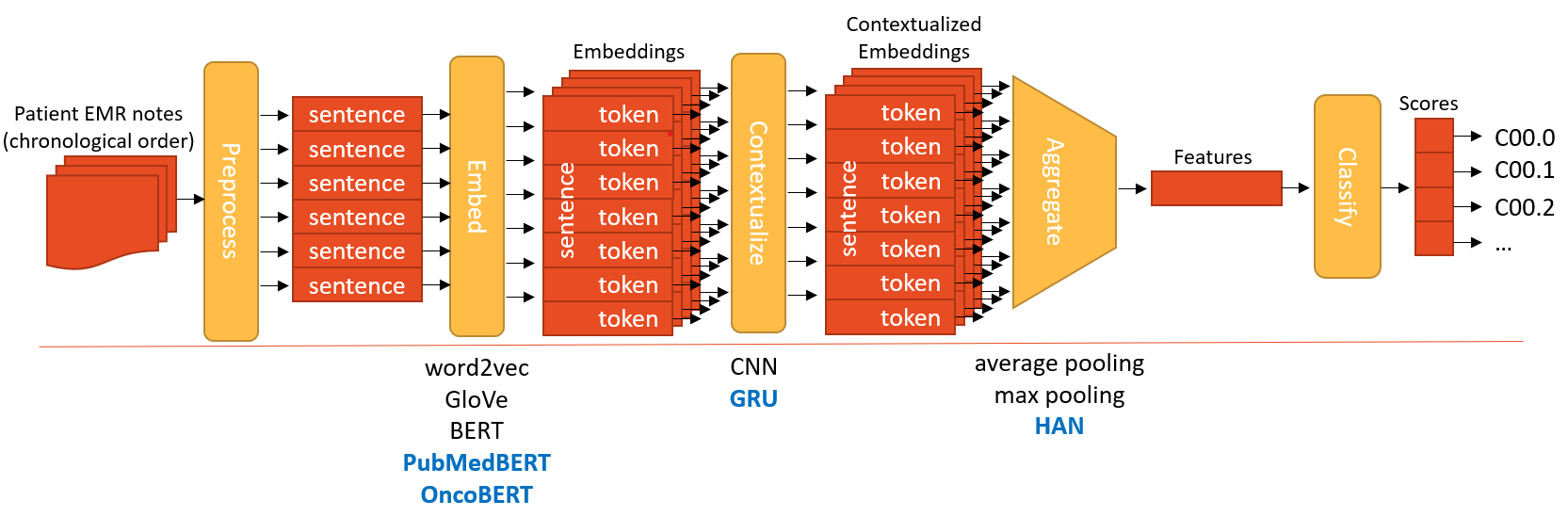}
    \caption{
    A general neural architecture for medical abstraction: clinical documents are concatenated by chronological order and converted into a token sequence, which is then transformed into a sequence of neural vectors by the embedding and contextualization modules, before being converted into a fixed-length feature vector by an aggregation module for final classification.
    }
    \label{fig:architecture}
\end{figure}

\begin{figure}[t]
    \centering
    \includegraphics[width=0.7\textwidth]{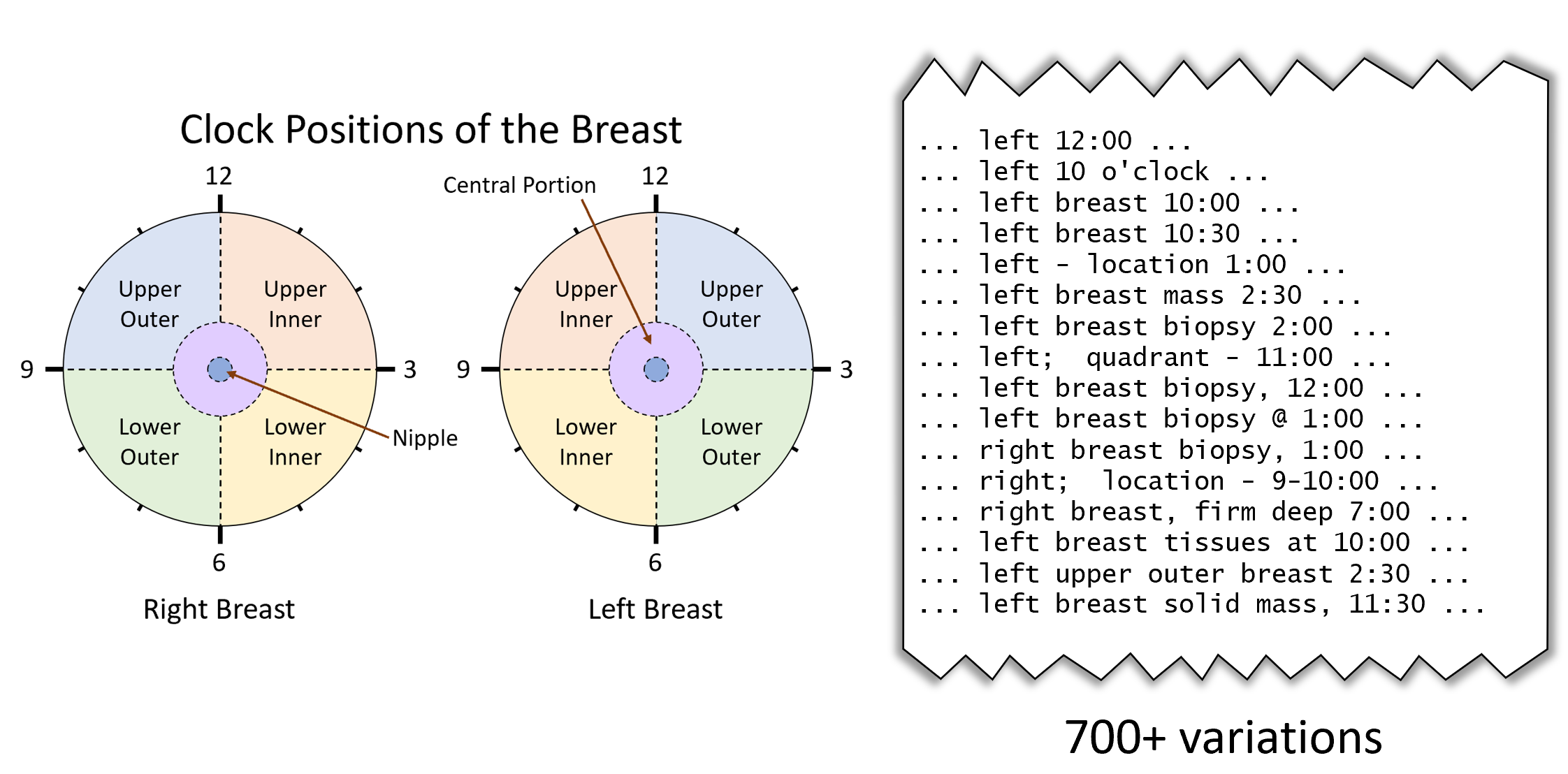}
    \caption{
    Relevant information for medical abstraction may manifest in myriad variations, as seen in specification of tumor site in breast cancer with laterality and clockwise postion.
    }
    \label{fig:variations}
\end{figure}

\begin{figure}[t]
    \centering
    \begin{tabular}{cc}
    \includegraphics[height=2.5in]{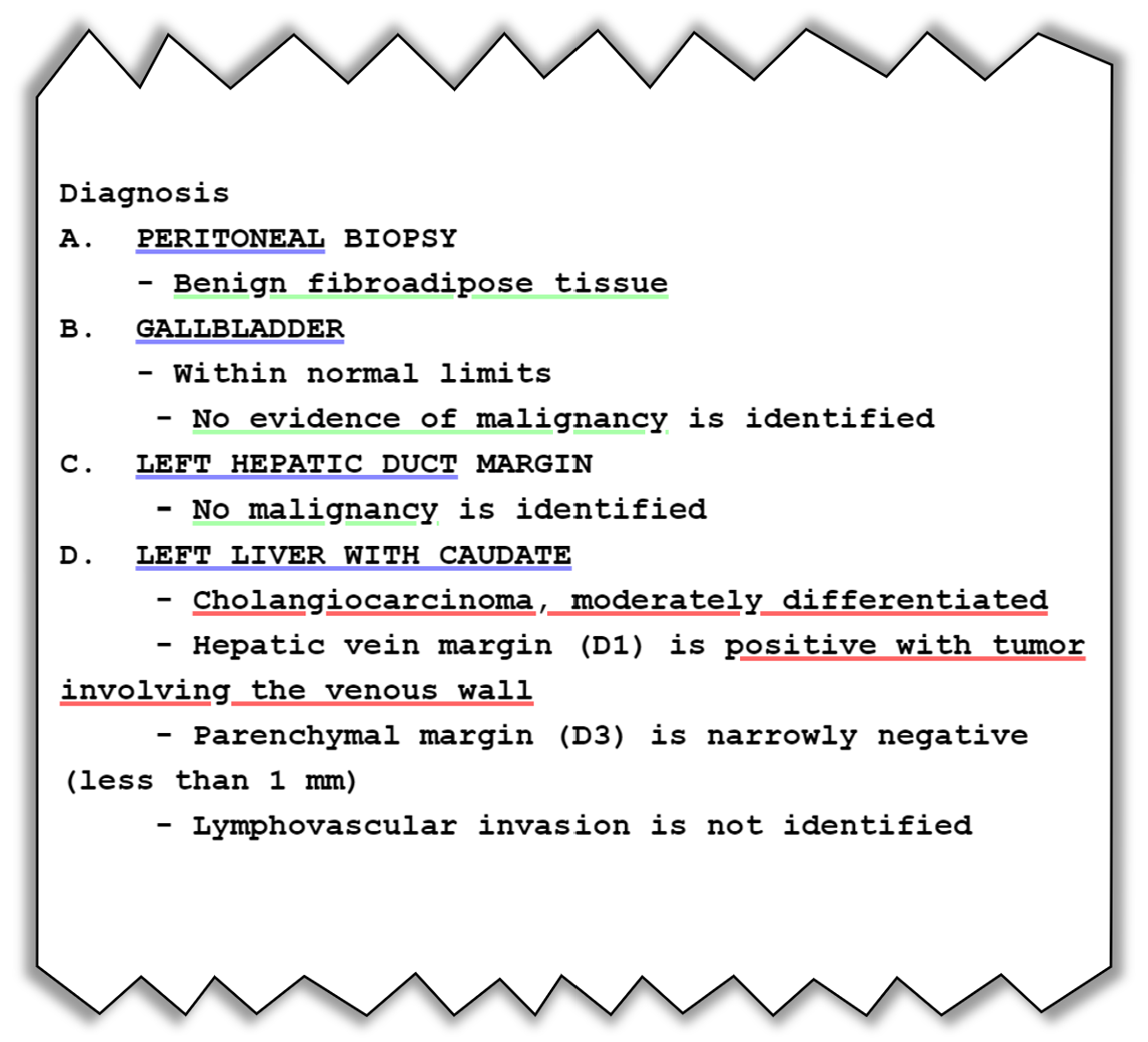} &
    \includegraphics[height=2.5in]{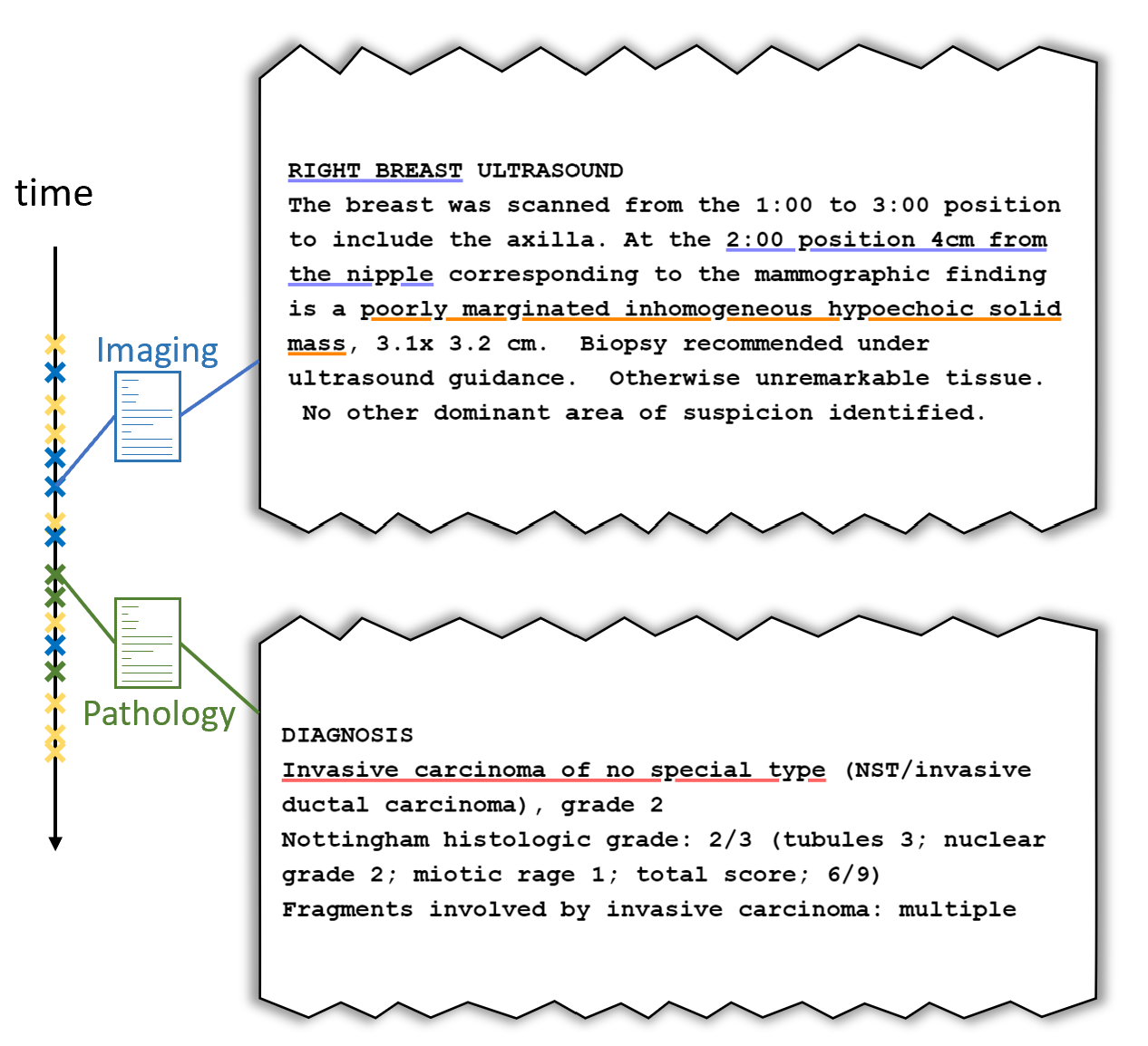} \\
    a & b
    \end{tabular}
    \caption{
    Simple NLP methods are not sufficient to handle complex semantics in general medical abstraction, as can be seen in the example of tumor site extraction. (a) NER is not enough; many candidate sites may be present, but the correct tumor site must be associated with a positive diagnosis. (b) Abstraction may require cross-document extraction; in this example the location is described in an imaging report, whereas the positive diagnosis is documented in a pathology report. Our sophisticated transformer-based model can classify them correctly and identify relevant rationale via attention. In these examples, blue underlining shows body sites, and green, orange, and red underlining show indications of negative, possible, and positive cancer diagnoses, respectively.
    }
    \label{fig:nlp_challenges}
\end{figure}

\eat{

\begin{figure}[t]
    \centering
    \begin{tabular}{ccc}
    \includegraphics[height=1.4in]{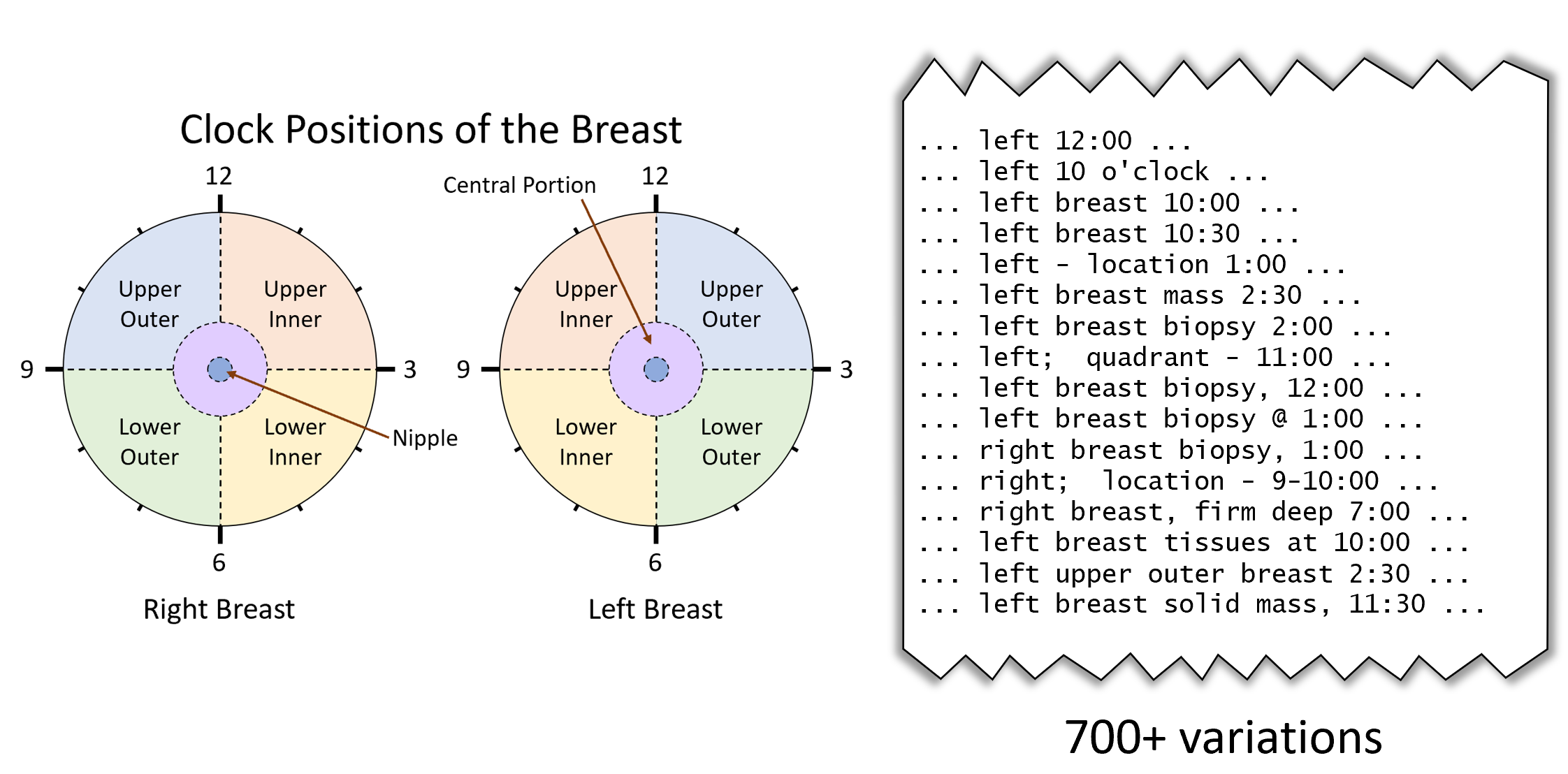} &
    \includegraphics[height=1.4in]{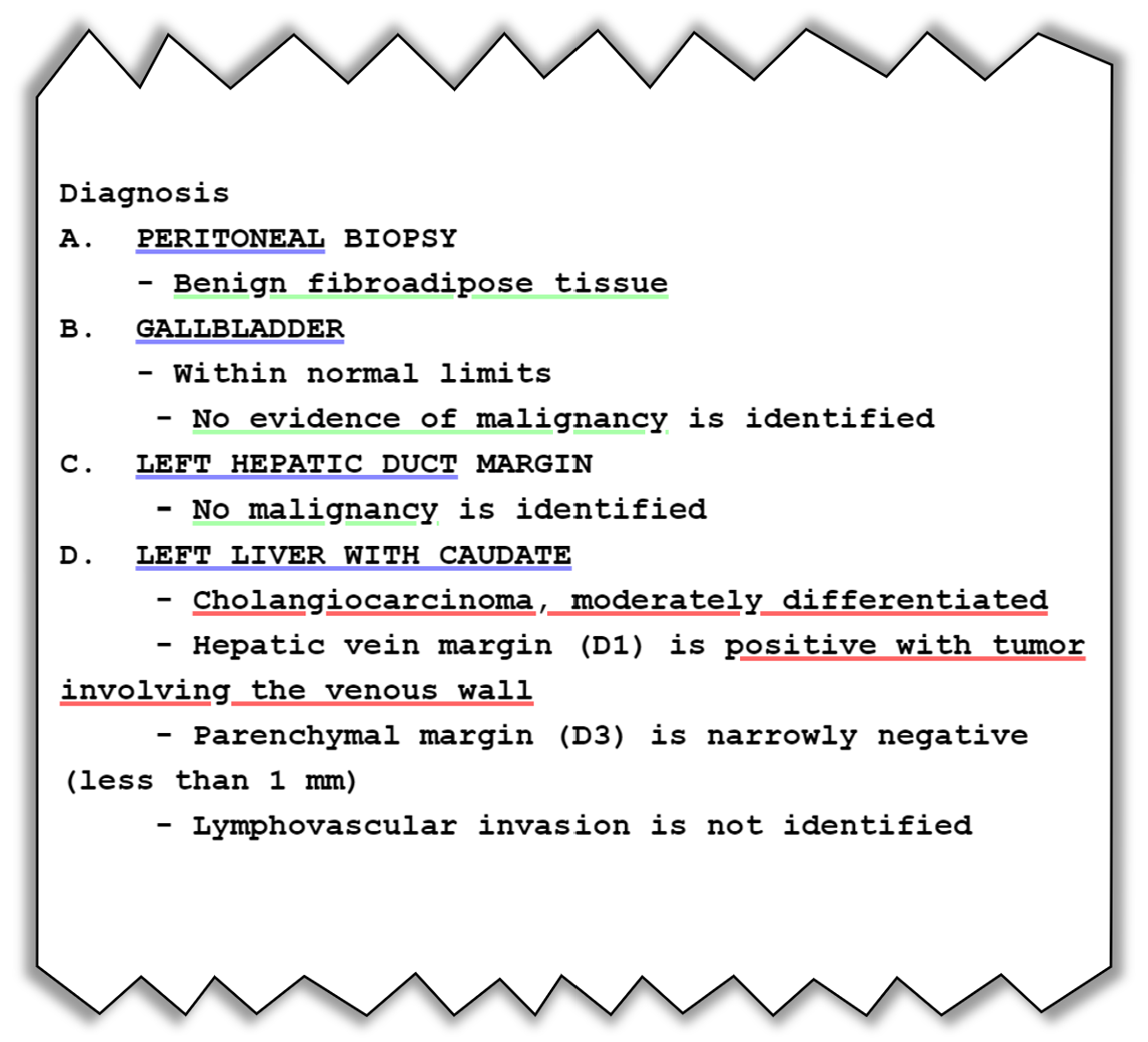} &
    \includegraphics[height=1.4in]{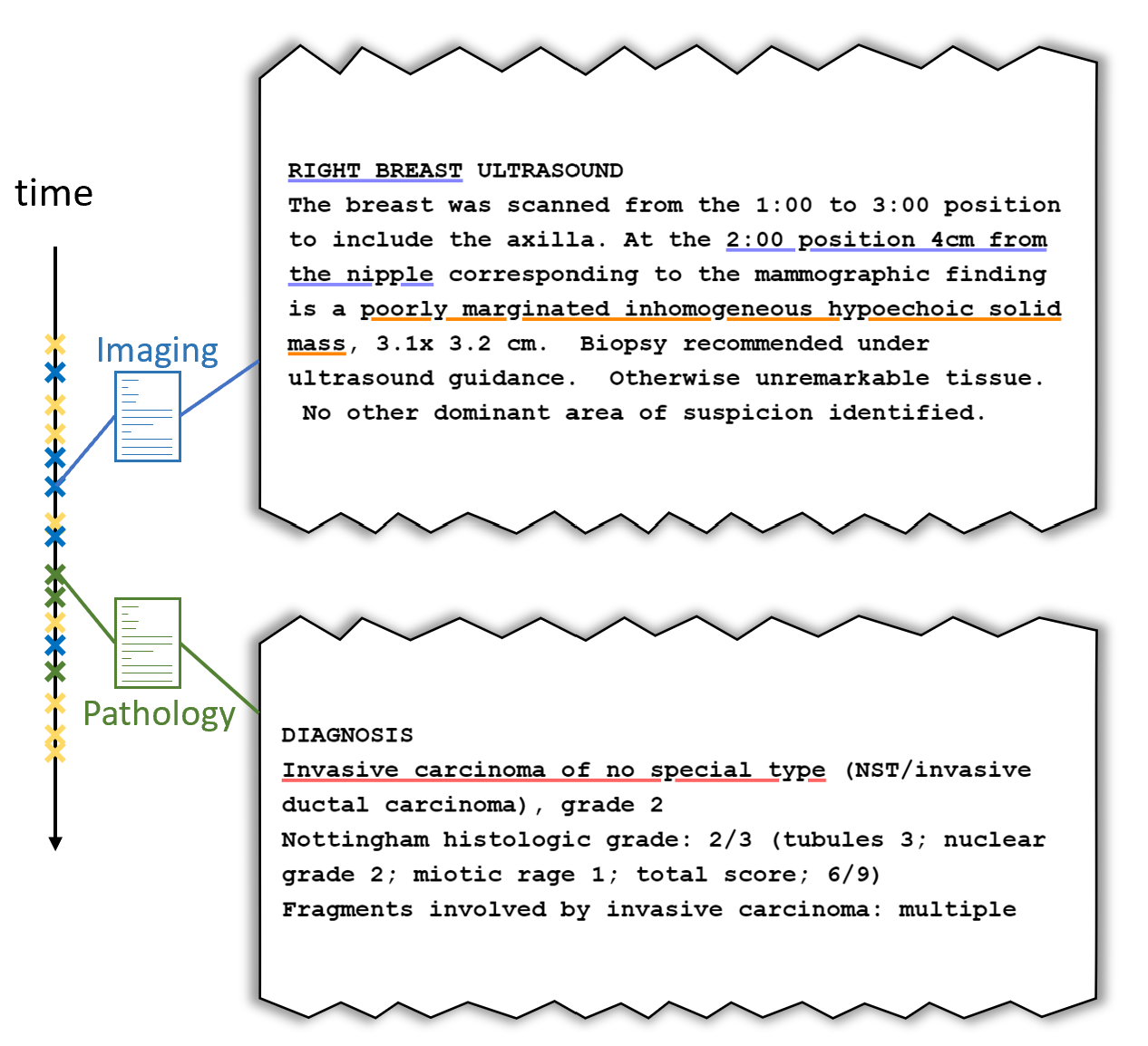} \\
    a & b & c
    \end{tabular}
    \caption{
    Phenotyping models face a variety of linguistic challenges, even in identifying tumor sites. (a) shows the clock-based language often used to identify the location of breast tumors, which requires both laterality and clock postion, and can be described using a wide variety of language. (b) shows how simple body site identification is inadequate, because many sites are described, and the correct location must be associated with a positive tumor diagnosis. (c) shows how even single-note context is not enough, because the precise location may only be described in imaging reports, while positive diagnosis is documented in pathology reports.
    }
    \label{fig:nlp_challenges}
\end{figure}
}


\subsection{Abstraction}

Medical abstraction can be formulated as information extraction in NLP. Given clinical text $T$ for a patient and attribute $A$, the goal is to extract $A$'s value as described in $T$ (or absence thereof), which can be framed as a multiclass classification problem. 
In most prior work, $T$ is a sentence or a clinical note, and $A$'s value only has a few choices (e.g., presence or absence of active cancer~\cite{kehl2019assessment}). By contrast, we consider the most general setting, where $T$ comprises all notes for a patient and $A$'s range may number in hundreds. E.g., there are 310 classes for tumor site and 556 for histology in ICD-O-3, and a patient may have many notes  (Figure~\ref{fig:multiple-notes}). 

In general, abstraction presents substantial challenges for NLP systems. Relevant information may manifest in myriad variations (Figure~\ref{fig:variations}). Name entity recognition (NER) is not enough, as abstraction is more about extracting underlying relations. E.g., abstracting tumor site is not about recognizing site mentions, but to determine if the patient has malignancy at the given site on a given date (Figure~\ref{fig:nlp_challenges}, a). Moreover, abstraction may require information integration across multiple clinical documents (Figure~\ref{fig:nlp_challenges}, b). 

With patient-level supervision from medical registries, our machine learning setting can be regarded as a form of distant supervision or more generally self-supervision~\cite{zhu21distantly}, as the labels cannot be attributed to a sentence or even a clinical document. However, given the aforementioned complex linguistic phenomena in medical abstraction, we do not generate noisy training examples by associating a label with a specific text span (e.g., individual sentences with the presence of relevant entities), as in standard distant supervision. Instead, we combine all clinical documents for a patient as input and rely on the deep-learning method to automatically identify pertinent sentences and notes.

\subsection{Deep Learning for Medical Abstraction}

Traditional clinical NLP systems are often rule-based, e.g., by leveraging regular expressions and domain lexicons from ontologies~\cite{meystre&al22}. They require significant efforts to build and may be vulnerable to linguistic variations and ambiguities. Consequently, machine-learning methods have seen increasing adoption~\cite{yim&al16}. Traditional learning-based NLP methods require users to provide feature templates for classification, whereas modern deep learning methods forgo this requirement and can automatically transform input text into a neural feature representation (a real-number vector)~\cite{wu&al20,oliver&al21,gao2021limitations,gao2019classifying,kehl2019assessment}. 


\autoref{fig:architecture} shows a general deep learning architecture for medical abstraction. Medical documents are ordered temporally and converted into a sequence of sentences. They are tokenized and converted into a neural representation by an embedding module where each token is turned into a real-number vector. The vectors are then updated by a contextualization module and combined into a fixed-length feature vector by an aggregation module, which the classification module uses as input to produce the final classification.

In prior work applying deep learning to medical abstraction, the embedding module generally uses simple context-free embedding such as \textit{word2vec}~\cite{mikolov2013distributed} or \textit{GLoVE}~\cite{pennington2014glove}. 
Contextualization is usually done by convolutional neural network (CNN), which runs a sliding window over the tokens and generates output vectors using a shared neural network, with aggregation done by pooling. 

Recently, there has been substantial progress in deep NLP methods. Transformer~\cite{vaswani2017attention}, with its multi-layer, multi-head self-attention mechanism, has proven very effective in modeling long-range dependencies and leveraging GPU parallelism. Contextualized embedding from language model pretraining~\cite{devlin2019bert,peters-etal-2018-deep} is much more powerful than context-free embedding such as Word2Vec and GLoVe in extracting semantic information from unlabeled text and modeling variations/ambiguities. While the bulk of pretraining work focuses on general domains such as newswire and Web, domain-specific pretraining has proven beneficial for specialized domains such as biomedicine, by prioritizing learning of biomedical terms in relevant biomedical contexts~\cite{peng2019transfer,lee2020biobert,gu2020domain}. 

In this paper, we conduct a thorough study of advanced deep NLP techniques in medical abstraction (\autoref{fig:architecture}, blue). 
Some prior work investigated deep NLP in simplistic settings (e.g., classifying individual pathology reports) and concluded that advanced techniques such as transformer does not help their tasks~\cite{gao2021limitations}. By contrast, we find that in the real-world setting of cross-document medical abstraction, advanced NLP techniques can confer significant benefit in combating the prevalent noises and linguistic complexities. 


For embedding, we use the state-of-the-art biomedical neural language model PubMedBERT~\cite{gu2020domain}. The input to a neural language model consists of text spans, such as sentences, separated by special tokens~$\tt [SEP]$. To address the problem of out-of-vocabulary words, neural language models generate a vocabulary from subword units~\cite{kudo2018sentencepiece,sennrich2015bpe}, by greedily identifying a small set of subwords that can compactly form all words in a given corpus. BERT~\citep{devlin2019bert} is a state-of-the-art language model 
based on transformer~\citep{vaswani2017attention}, which is pretrained by predicting held-out words in unlabeled text. 
While most BERT models were pretrained on general-domain text~\citep{devlin2019bert,liu2019roberta}, PubMedBERT instead uses a biomedicine-specific vocabulary and was pretraining on biomedical literature from scratch. 
We also pretrain an oncology-specific OncoBERT on EMRs from over one million cancer patients and explore its use in oncology abstraction. 

Self-attention requires pairwise computation among tokens, which scales quadratically in input text length. Consequently, standard BERT models typically limit input length (e.g., 512 tokens). This is not a problem for restricted settings such as sentence-level or document-level abstraction in prior work, but it poses a substantial challenge in the general setting, as patient-level cross-document input has a median length of over 4,000 tokens. To handle such long text, we use gated recurrent unit (GRU)~\cite{cho2014learning} for contextualization and hierarchical attention network (HAN)~\cite{yang2016hierarchical} for aggregation. GRU helps propagates information beyond BERT’s default length limit and HAN provides better aggregation than pooling by weighing relevant tokens higher. The classification module is a standard linear layer followed by $\tt softmax$, which produces multinomial probabilities among possible labels.

\subsection{Case Finding}


In medical abstraction, we are given cancer patients and asked to extract key tumor attributes. By contrast, the goal of case finding is to determine if a patient should be included for cancer registry. This can be framed as binary classification over a patient’s clinical documents from a given day. We use the same architecture as in \autoref{fig:architecture} and find it similarly effective. (The models are learned separately for case finding vs. abstraction. We conducted preliminary experiments on multi-task learning but didn't find significant difference in performance, as each task has abundant training data.)

Case finding poses a distinct self-supervision challenge. We can easily identify positive examples from the registry (patients with their diagnosis dates). However, it is less clear how to identify negative examples. We explore two self-supervision schemes. Initially, we randomly sample non-cancer patients and days from their medical history with pathology reports. This yields a classifier with good sensitivity (recall), but often incorrectly flags pre-diagnosis days for a cancer patient, causing a high false-positive rate. To address this problem, we experiment with adding hard negative examples from cancer patients by sampling days before diagnosis. The resulting classifier not only distinguishes cancer patients from non-cancer patients, but also identifies the time of initial diagnosis, as required for case finding. Together with abstraction, we can thus help accelerate cancer registry curation end-to-end.

\subsection{Human Subjects / IRB, Data Security \& Patient Privacy}

This work was performed under the auspices of an Institutional Review Board (IRB)-approved research protocol (Providence protocol ID 2019000204) and was conducted in compliance with Human Subjects research and clinical data management procedures---as well as cloud information security policies and controls---administered within Providence St. Joseph Health.  All study data were integrated, managed and analyzed exclusively and solely on Providence-managed cloud infrastructure.  All study personnel completed and were credentialed in training modules covering Human Subjects research, use of clinical data in research, and appropriate use of IT resources and IRB-approved data assets. 

\section{Results}


We conduct experiments using data from a large integrated delivery network (IDN) with over 28 distinct cancer care centers across U.S. states.  
We assemble a dataset with patient-level supervision by matching comprehensive EMR records (including all free-text clinical documents in scope here) and cancer registry records. Patients without a digitized pathology report within 30 days of diagnosis are skipped. This yields a total of 135,107 patients spanning multiple U.S. states, between 2000-2020. We use patients in Oregon for the initial exploration ($n=39,064$, 29\% of patients).
We divide patients into ten random folds. 
We use six folds for training and development ($n=23,438$), two folds for test ($n=7,740$), and two folds for additional held-out test set ($n=7,881$). We reserve patients from Washington ($n=36,900$), as well as the remaining states ($n=59,143$) for further generalizability test, with a distinct health system being used in each state.


\eat{
\begin{table}[ht]
\begin{center}
\begin{tabular}{ccccc}
& Oregon Train & Ore & Washington & Other States \\
\midrule
Patient# &  &  & 69.1 \\
Histology & 99.4 & 87.2 & 81.2\\
Clinical T & 93.9 & 79.3 & 70.1\\
Clinical N & 97.2 & 97.2 & 91.6\\
Clinical M & 98.7 & 98.7 & 94.9\\
Pathological T & 96.1 & 87.2 & 78.6\\
Pathological N & 95.2 & 95.3 & 88.9\\
Pathological M & 98.6 & 98.6 & 95.1\\
\bottomrule
\end{tabular}
\end{center}
\caption{
Patients used in our evaluation.
}
\label{tab:dataset}
\end{table}
}


We use the ICD-O-3 ontology for tumor site and histology. For staging, we focus on solid tumors and follow AJCC guidelines for clinical and pathological staging. Both represent cancer progression using TNM classification (T: tumor size/location; N: lymph node status; M: metastasis). Clinical staging is based on initial diagnosis using medical imaging, clinical assessment and/or biopsy, whereas pathological staging incorporates more definitive assessment of the tumor size and spread. For simplicity and based on practical utility, we focus on classifying coarse categories (T: 0-4, in situ; N: 0 vs 1+; M: 0 vs 1).  


For each attribute, we report standard area under the receiver operating characteristic curve (AUROC). For system comparison, however, AUROC might obscure key performance difference in the presence of imbalanced distribution (e.g., some sites appear much more frequently), so we evaluate area under the precision-recall curve (AUPRC). We also report accuracy for completeness. Precision and recall are also known as positive predictive value and sensitivity, respectively.

\subsection{Main Results}

\begin{table}[ht]
\begin{center}
\begin{tabular}{lccc}
& AUROC & AUPRC & accuracy \\
\midrule
Tumor Site & 99.3 & 76.7 & 69.1 \\
Histology & 99.4 & 87.2 & 81.2\\
Clinical T & 93.9 & 79.3 & 70.1\\
Clinical N & 97.2 & 97.2 & 91.6\\
Clinical M & 98.7 & 98.7 & 94.9\\
Pathological T & 96.1 & 87.2 & 78.6\\
Pathological N & 95.2 & 95.3 & 88.9\\
Pathological M & 98.6 & 98.6 & 95.1\\
\bottomrule
\end{tabular}
\end{center}
\caption{
Test results for oncology abstraction by our deep learning system based on PubMedBERT.
}
\label{tab:summary-results}
\end{table}

\autoref{tab:summary-results} shows test results for extracting key oncology attributes. By incorporating state-of-the-art advances such as PubMedBERT, our deep learning system attains high performance across the board, even for tumor site and histology where the system has to distinguish among hundreds of labels. Despite the large parameter space, our system is robust in experiments, with standard deviation across two random runs smaller than 1\% for all tasks.

\subsection{Generalizability}

\begin{table}[ht]
\begin{center}
\begin{tabular}{lcccc}
& OR Test & OR Held-Out & WA & Other States \\
\midrule
Tumor Site & 76.7 & 76.4 & 73.5 & 73.0 \\
Histology & 87.2 & 87.6 & 80.5 & 78.0\\
Clinical T & 79.3 & 78.8 & 73.5 & 73.5\\
Clinical N & 97.2 & 97.6 & 95.4 & 96.0\\
Clinical M & 98.7 & 98.8 & 97.3 & 97.7\\
Pathological T & 87.2 & 88.0 & 84.3 & 86.1\\
Pathological N & 95.3 & 95.7 & 92.9 & 95.1\\
Pathological M & 98.6 & 98.6 & 97.1 & 97.1\\
\bottomrule
\end{tabular}
\end{center}
\caption{
Generalizability test: all results were obtained using our deep-learning models (based on PubMedBERT) trained on Oregon Training. Washington (WA) and other states all use different health systems. There is only slight degradation for most results, which bodes well for generalizability of our models. A notable exception is histology, with up to 9 points drop. Upon close inspection, this stems from divergence in curation standards on ambiguous cases, with registrars using different labeling granularity (e.g., non small-cell lung cancer vs. lung adenocarcinoma).
}
\label{tab:generalizability}
\end{table}

To assess the generalizability, we evaluate on the held-out set and find that model performance is nearly identical. We further evaluate our model trained on the Oregon training set on patients from Washington and other states. See \autoref{tab:generalizability}. The results are comparable for most attributes, with only slight degradation. Histology, however, shows a large performance decrease (87.2 to 80.5 and 78.0). Manual analysis shows that much of this drop is attributable to differences in curation standards, with registrars from different systems using different labeling granularity, e.g., non small cell lung cancer (8046) vs. adenocarcinoma (8140), with the latter being the most common type of the former.
Clinical T also shows a noticeable performance decrease (79.3 to 73.5). Manually analysis shows that this performance drop largely stems from higher proportion of highly ambiguous cases (e.g., borderline categories between Stage 2 and 3).


\subsection{System Comparison}

\begin{table}[ht]
\begin{center}
\begin{tabular}{lcccccccc}
& Tumor Site & Histology & Clinical T & N & M & Pathological T & N & M \\
\midrule
Ontology & 19.4 & 19.2 & - & - & - & - & - & - \\
BOW & 62.8 & 76.6 & 70.4 & 96.6 & 98.4 & 72.1 & 90.7 & \textbf{98.9} \\
OncoGloVe + CNN & 72.0 & 84.4 & 74.2 & 96.5 & 98.6 & 83.9 & 93.1 & 98.5\\
OncoGloVe + HAN/GRU & 74.0 & 85.9 & 76.2 & 97.1 & 98.7 & 86.4 & 94.2 & 98.5\\
BERT + HAN/GRU & 75.1 & 86.2 & 77.0 & 96.6 & 98.4 & 86.4 & 94.4 & 98.2\\
PubMedBERT + HAN/GRU (ours) & 76.7 & 87.2 & 79.3 & 97.2 & 98.7 & 87.2 & 95.2 & 98.6 \\
OncoBERT + HAN/GRU (ours) & \textbf{77.1} & \textbf{87.6} & \textbf{81.4} & \textbf{97.5} & \textbf{99.0} & \textbf{87.6} & \textbf{95.5} & \textbf{98.9} \\
\bottomrule
\end{tabular}
\end{center}
\caption{
Comparison of test AUPRC scores for oncology abstraction by various NLP systems. Ontology: ontology-aware rule-based system. BOW: logistic regression with bag-of-word features. OncoGloVe: 100-dimensional GloVe embedding pretrained on oncology notes.
}
\label{tab:sys-comparison}
\end{table}

\autoref{tab:sys-comparison} compares our deep learning systems with prior approaches for medical abstraction. An ontology-aware rule-based system (matching against class lexicon and known aliases) performs poorly, demonstrating that entity recognition alone is inadequate for such challenging tasks. Deep-learning methods perform substantially better, with BERT-based approaches outperforming CNN, especially for the most challenging tasks such as site, histology, and clinical/pathological tumor (T) staging. HAN/GRU and transformer-based language models each contribute significantly, with our best system gaining 5.1 points for site, 3.2 points for histology, and 7.2 points for clinical T over GloVe+CNN. 

Domain-specific pretraining is especially impactful. By pretraining entirely on oncology notes, OncoBERT further improves over PubMedBERT, which is already pretrained on biomedical text. Compared to general-domain BERT, our best system with OncoBERT gains 2.0 points for site and 4.4 points for clinical tumor (T) staging.  


\subsection{Ablation Study}

We incorporate three types of clinical documents as input: pathology report, radiology reports, and operative notes. In ablation study, we find that having all three helps, presumably because this increases robustness in case some relevant notes are missing or not yet digitized (e.g., scanned PDFs). E.g., adding radiology reports on top of pathology reports increased AUPRC by 3.4 absolute points for tumor-site extraction, with the inclusion of operative notes providing an additional one point gain. By default, we use [-30, 30] days around diagnosis, which works reasonably well in general. For pathological staging, however, a larger window is helpful, as relevant information often comes from a tumor surgical resection that may be several months after an initial tissue biopsy or fine needle aspiration based diagnosis. E.g., using [-30, 90] days as input improves AUPRC by over four absolute points for pathological tumor (T) staging (87.2 vs 91.8).

\subsection{Case Finding}

\begin{table}[ht]
\begin{center}
\begin{tabular}{cccc}
Self-Supervision & Train Positive Instances & Train Negative Instances & Test F1 \\
\midrule
Default & 37,207 & 13,123 & 91.4\\
+ Hard Negatives & 37,207 & 22,959 & 97.3\\
\bottomrule
\end{tabular}
\end{center}
\caption{
Comparison of test results in case finding with two self-supervision schemes.
}
\label{tab:case-finding}
\end{table}


We assemble a case-finding dataset using patients in the cancer registry. For positive cases, we identify cancer patients with at least a pathology report on the day of diagnosis. For negative cases, we randomly sample non-cancer patients. This yields 62,090 positive and 8,460 negative patients. We divide them into train/development/test by 60\%/20\%/20\%, with 12,418 positive and 1,692 negative patients in the test set.

A patient may have clinical documents in multiple days. In case finding, a classification instance comprises a patient's clinical documents in a given day, and the ultimate goal is to identify the moment of cancer diagnosis (when registry curation starts). For evaluation, we adopt a patient-level metric that mirrors real-world applications. For each patient, we return the first day with positive classification. For cancer patients, the case-finding decision is deemed correct if the first day of positive classification is within [-7, 30] days of diagnosis. For non-cancer patients, the case-finding decision is deemed correct if all classifications are negative. 
We report the F1 score which is the harmonic mean of precision (positive predictive value) and recall (sensitivity). Specifically, $F1=2/(1/precision + 1/recall)$.


For self-supervision, we explore the two settings as described in the Method Section. In both cases, positive instances comprise cancer patients on the diagnosis date. By default, negative instances comprise of randomly chosen days among non-cancer patients. Additionally, we randomly sample days before diagnosis among cancer patients and add 9,836 instances as hard negative examples. 
With the base setting, we attain test F1 91.4. By incorporating hard negative examples, we substantially improve test F1 to 97.3, gaining six absolute points.

\section{DISCUSSION AND CONCLUSION}

\subsection{Error Analysis}
We conduct manual analysis on sample errors. Some stem from annotation inconsistency, where registrars actually agree with our system classifications upon close inspection. Others stem from missing notes. After adjusting for annotation inconsistency and missing input, the real test performance of our deep-learning system is even higher. For example, based on estimate from analyzing fifty error examples, the real test AUPRC for tumor site is about 91.6 (vs. 76.7).  

\subsection{Assisted Curation}

\begin{figure}[t]
    \centering
    \begin{tabular}{cc}
        \begin{tabular}{c}
            \includegraphics[width=2.5in]{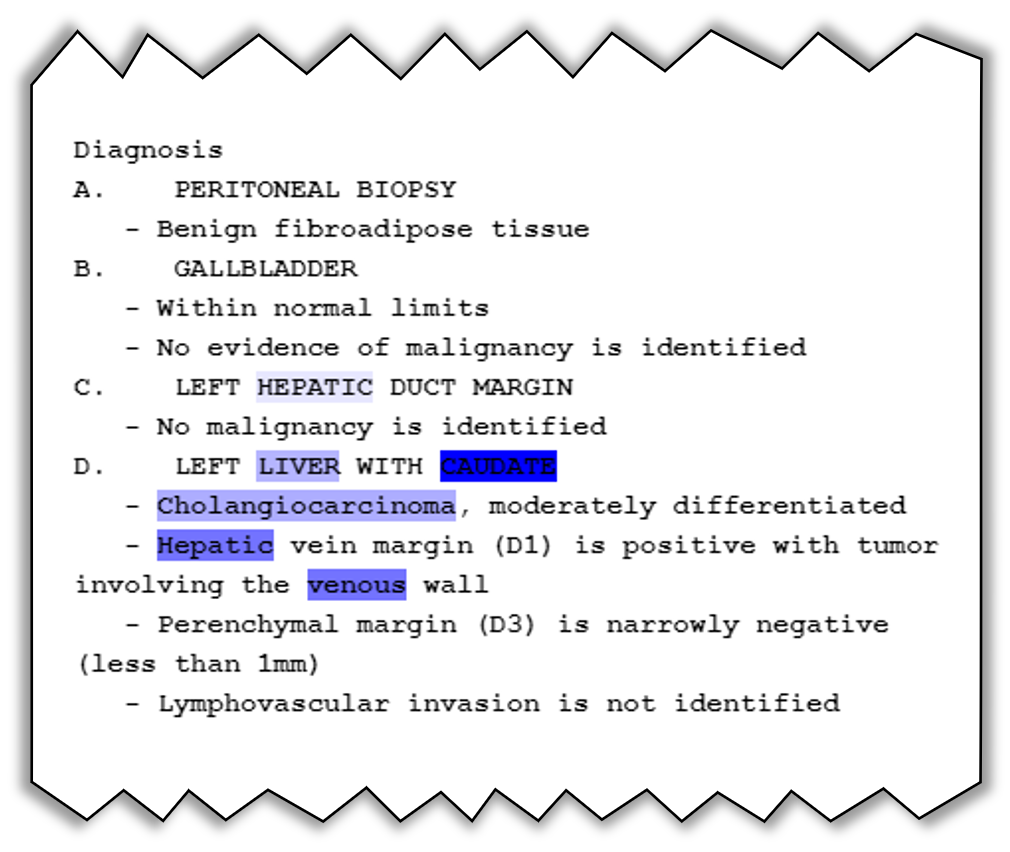} \\
            (a) C22.1: Intrahepatic bile duct
        \end{tabular}
        \begin{tabular}{c}
            \includegraphics[width=2.5in]{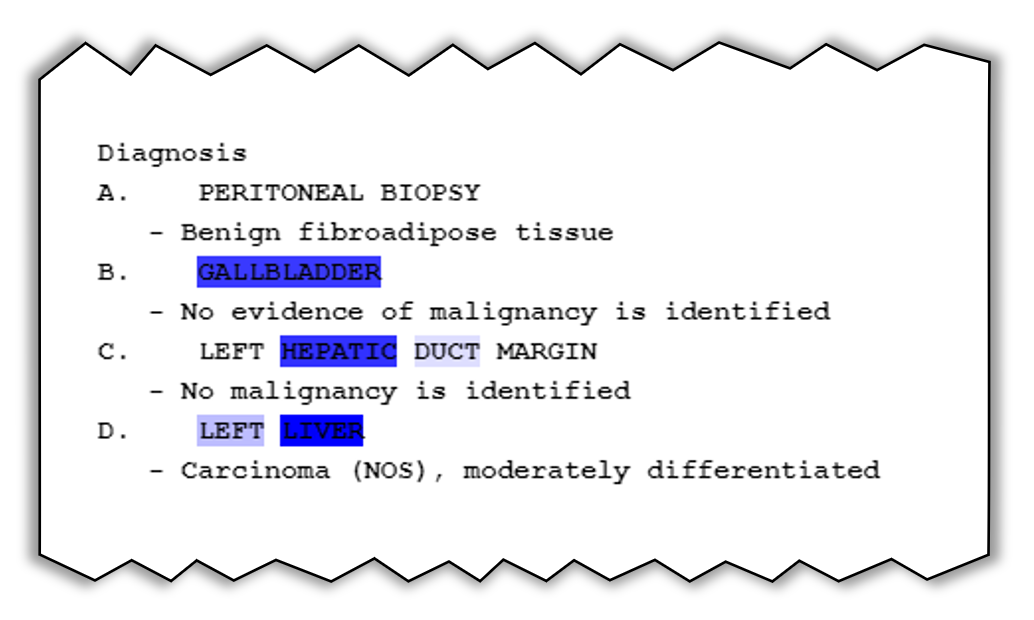} \\
            (b) C22.0: Liver\\
            \includegraphics[width=2.5in]{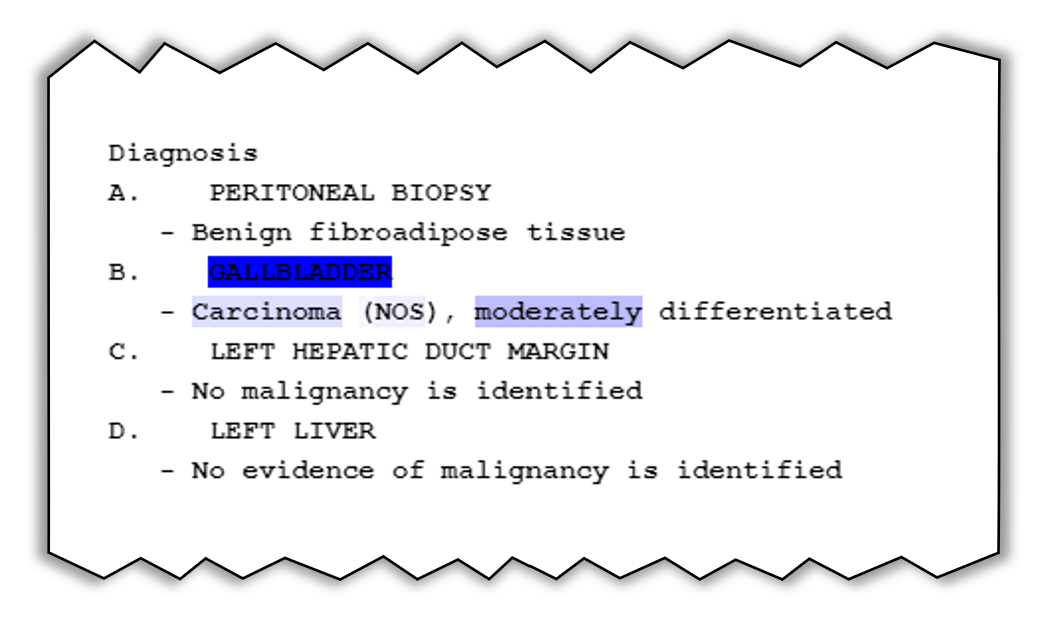} \\
            (c) C23.9: Gallbladder \\
        \end{tabular}
    \end{tabular}
    \caption{Examples of observed attention patterns and predictions from the tumor site model. (a) shows the attention pattern for the example shown in \autoref{fig:nlp_challenges} a, with darker color signifying higher attention weight. The tumor site model correctly identifies \emph{C22.1: Intrahepatic bile duct} due to the cholangiocarinoma histology (indicating cancer of the bile duct). To probe the model understanding further, inference was run on modified text. In (b), the description was changed to a generic ``carcinoma'' diagnosis. While the attention is more diffuse, the model places the highest attention on the ``liver" section, and correctly identifies \emph{C22.0: Liver} as the tumor site. In (c), the ``carcinoma'' diagnosis was moved to the ``gallbladder'' section, and the model now correctly identifies the site as \emph{C23.9: Gallbladder}, with attentions now focusing on this section.
    }
    \label{fig:attention_examples}
\end{figure}

\begin{figure}
    \centering
    \includegraphics[width=0.9\textwidth]{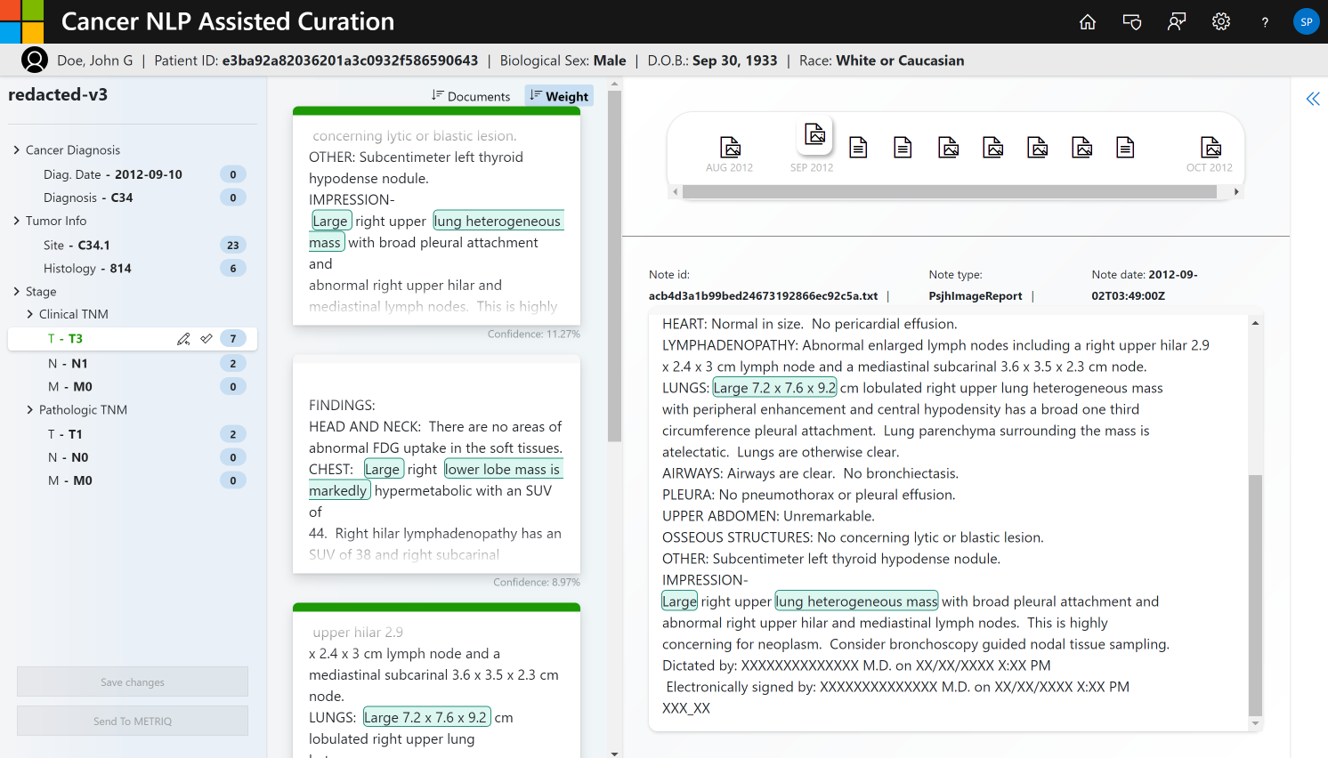}
    \caption{
    Our cancer assisted curation system. Left: extracted oncology attributes. Middle: extraction rationale based on attention weights. Right: full notes. Patient information has been de-identified. 
    }
    \label{fig:assisted-curation}
\end{figure}

We envision that NLP extraction can serve as candidates to help accelerate curation. 
The attention mechanism in transformer-based model provides a straightforward approach to identify extraction rationale. Effectively, the aggregate representation of the input text is the weighted sum of token representations in the top layer, with the weights (derived from self-attention to the special $\tt [CLS]$ token) signifying relative importance of individual tokens in the final classification decision. While there is no guarantee that attentions provide explanation~\cite{jain&wallace19}, in practice we find that tokens with the highest attention weights are conducive to assisted curation and generally conform with what human experts would consider as extraction rationale.
As an example, \autoref{fig:attention_examples} highlights tokens with high attention weights for the example text in \autoref{fig:nlp_challenges}a and two variations. While the attention may not entirely align with individual human intuition, it broadly conforms with the extraction rationale and enables quick verification. 
\autoref{fig:assisted-curation} shows a prototype we have developed for assisted curation, which is in beta test by selected clinical users. For each attribute, the interface displays the extraction rationale by highlighting individual note and text span with the highest neural attention weight for final classification. In preliminary studies, tumor registrars can verify a candidate extraction in 1-2 minutes, either ascertaining its correctness or fixing the label in the interface.  

\subsection{Limitations}

Our study focuses on medical abstraction of key diagnosis information as curated in cancer registry. Future work should explore extraction of treatment and outcome information, as well as other diagnostic information such as biomarkers. 
Cancer registry focuses on complete curation of ``analytical cases'', i.e., patients with both initial diagnosis and treatment occurring within a given healthcare system. The models may perform less well for patients who are initially diagnosed elsewhere and then referred to the given network, e.g., due to missing digitized reports. In many such cases, PDFs or scanned documents are still available. We are exploring the use of state-of-the-art document image understanding methods, such as LayoutLM~\cite{xu&al20-layoutlm}, with initial promising results. 
Our immediate exploration of assisted curation focuses on accelerating case identification and medical abstraction, but it also opens up opportunities for interactive learning to continuously improve machine reading based on user feedback.  
In addition to improving abstraction accuracy, this can potentially help calibrate attention weights for extraction rationale~\cite{bao&al18supervisedrationale}. 
Pretraining can also be further improved by incorporating domain knowledge such as UMLS~\cite{hao-etal-2020-enhancing,sheng&al21}. 

\subsection{Towards Scaling Real-World Data Curation}


Manual curation of complex clinical records and EHR data is expensive and time-consuming. The healthcare network represented in this study hires several dozens of full-time registrars for cancer registry abstraction. Curation is limited to analytic cases (i.e. those first treated in a given cancer center), which are legally required for reporting, thus skipping a large swath of patients. Despite such restrictions and significant investment, there is still significant delay for majority of the patients. To estimate the extent of curation backlog, we analyze two snapshots of cancer registry that are eight months apart. Among newly curated cases in the second snapshot, 23,670 are diagnosed before the first snapshot ends. They have a median of 324 days between diagnosis and the first snapshot end date. Many cases are curated over a year after diagnosis. By leveraging assisted curation with candidate abstractions generated by our deep NLP system, we can accelerate cancer registry abstraction and reduce backlog. Given promising results in the preliminary study, we are now exploring integration of assisted curation to the registry abstraction workflow.

NLP-based machine reading also helps scale real-world data curation. The healthcare network in our study has over 1.2 million cancer patients with digitized pathology reports within 30 days of diagnosis. However, only 135,107 of them have been curated in the cancer registry. 
By applying our NLP system to all patients, we instantly expand structured real-world data for the network by an order of magnitude. 
In future work, we plan to expand the scope of curation by applying self-supervised learning to extracting other key information for real-world evidence, such as treatments and key clinical outcomes~\cite{ratner2016data,wang&poon18-DPL,hunter&poon21-S4,sheng&al21-modular}. A particularly exciting research frontier lies in studying response to immunotherapy, such as check-point inhibitors (CPI). 
In preliminary study, we find that self-supervised NLP methods can immediately identify and abstract over an order of magnitude more CPI patients, compared to prior manual efforts that took many months.

\section{COMPETING INTEREST}
There is NO Competing Interest.




\bibliographystyle{plain}
\bibliography{references}

\end{document}